\documentclass[11pt]{article}

\usepackage[final]{acl}

\usepackage{times}
\usepackage{latexsym}

\usepackage[T1]{fontenc}

\usepackage[utf8]{inputenc}

\usepackage{microtype}

\usepackage{inconsolata}

\usepackage{graphicx}

\usepackage{linguex}
\usepackage{booktabs}
\usepackage{amsmath}
\usepackage{siunitx}
\usepackage{enumitem}
\usepackage[export]{adjustbox}

\title{Not Worth Mentioning? \\ A Pilot Study on Salient Proposition Annotation}

\author{Amir Zeldes, Katherine Conhaim, Lauren Levine\\
  Department of Linguistics \\
  Georgetown University \\
  \texttt{\{amir.zeldes,kc1512,lel76\}@georgetown.edu} } 

\begin{document}
\maketitle
\begin{abstract}
Despite a long tradition of work on extractive summarization, which by nature aims to recover the most important propositions in a text, little work has been done on operationalizing graded proposition salience in naturally occurring data. In this paper, we adopt graded summarization-based salience as a metric from previous work on Salient Entity Extraction (SEE) and adapt it to quantify proposition salience. We define the annotation task, apply it to a small multi-genre dataset, evaluate agreement and carry out a preliminary study of the relationship between our metric and notions of discourse unit centrality in discourse parsing following Rhetorical Structure Theory (RST).
\end{abstract}

\section{Introduction}

Judging how important a proposition is to the contents of a text or conversation is important for a range of tasks in information retrieval, summarization and text mining \cite{narayan-etal-2018-ranking,NguyenLe2024,DwivediEtAl2025-11365815}. In extractive summarization in particular, it is implied that the most important propositions in a document will be selected for inclusion in summaries. However, little work has been done to explain which propositions should be considered salient, and what the operationalization of salience should be, or how it could be measured.

By contrast, salience annotation for entities has been studied more extensively. While some papers have shown that manual annotation of salient entities exhibits low agreement \cite{dojchinovski-etal-2016-crowdsourced,Trani2018SELAU}, an alternative paradigm leveraging document summaries \cite{dunietz-gillick-2014-new} has been more successful in delivering a consistent operationalization by equating salience with \textbf{summary worthiness}: since it is difficult to summarize a document without mentioning its most salient entities, entities mentioned in a summary can be considered salient. 

Two main criticisms of this approach have been recently addressed by \citet{lin-zeldes-2025-gum}: 1. that summaries themselves are subjective, meaning different summaries will result in different salience annotations; and 2. that the resulting annotations are binary (mentioned/not-mentioned in the summary). \citeauthor{lin-zeldes-2025-gum} address this by obtaining five summaries for each document, which allows for a gradient metric (entities mentioned in 5 summaries are more salient than those mentioned in 2 or 3), and controls for summary subjectivity by obtaining multiple judgments.

In this paper, we aim to test the same methodology to the concept of proposition salience. Specifically, we will define a gradient salience metric for propositions based on \textbf{how many summaries mention a proposition}. With the data at hand, we will provide some first insights on characteristics of salient propositions, and in particular, determine the extent to which the metric correlates with an existing notion of discourse unit centrality in the tradition of RST discourse parsing. The main contributions of this paper are:

\begin{itemize}[itemsep=0.8pt]
    \item Definition and evaluation of a summary-based approach to graded proposition salience
    \item A pilot dataset of salient propositions aligned to summaries with additional annotations
    \item Analysis of the correlation between our approach and RST centrality
    \item An annotation interface for the task
\end{itemize}

\section{Previous work}

Much of the work on ranking proposition salience has been carried out in the context of summarization tasks \cite{Saggion2013}, specifically optimizing for selecting sentences that maximize NLG evaluation metrics \cite{nallapati2017summarunner,narayan-etal-2018-ranking}, such as ROUGE scores \cite{lin-2004-rouge}. Early work on sentence or paragraph importance in texts focused on direct human judgment and recall tasks testing the memorability of parts of a text in closed book scenarios \cite{TRABASSO1985595,wolf-gibson-2004-paragraph-word}. Subsequent work on extractive summarization focused on identifying the most relevant sentences for direct inclusion in summaries. Seminal work by \citet{nenkova-passonneau-2004-evaluating} showed that propositions included in multiple reference summaries should be prioritized and used in the evaluation of system summaries. Later studies shifted focus to supervised approaches including tree-based methods leveraging semantic annotations \cite{fang-etal-2016-proposition}, end-to-end representation learning in embedding spaces \cite{ChenEtAl2018}, and more recently using LLM prompting \cite{parmar-etal-2024-towards}. 

Studies using annotated data have relied either on direct annotation of important sentences \cite{liu-etal-2018-automatic}, or on their derivation from extractive summarization datasets for sentence selection \cite{cheng-lapata-2016-neural}, both of which suffer from instability. Some previous work, such as \cite{liu-etal-2018-automatic}, also attempted to use lexical overlap (lemmas) between documents and summaries to approximate binary salience for sentences, but did not carry out extensive manual evaluation or attempt to use multiple summaries. The most similar previous work to the present paper is \citet{lin-zeldes-2024-gumsley,lin-zeldes-2025-gum}, which uses a summary-based methodology to rank salient entities based on the number of summaries they appear in, including manual verification of alignments. Below we apply analogous methods to ranking salient propositions.

\section{Data}

The data used by \citet{lin-zeldes-2025-gum} comes from the freely available GUM corpus \cite{Zeldes2017b}, which includes five summaries per document. These were collected in an earlier project aiming to produce consistent and faithful summaries across a broad range of genres using controlled guidelines, which instructed summarizers to remain as close as possible to the text \cite{liu-zeldes-2023-gumsum}, thereby facilitating our task. We re-use that data for this pilot study, annotating the test set, which comprises 32 documents from 16 genres of spoken and written English with \textasciitilde30K tokens and 3,800 propositions (see Table \ref{tab:data}). Coupled with 5 summaries per document (and therefore per proposition), our data consists of close to 19K proposition annotations (alignments).

\begin{table}[h!tb]
\resizebox{\columnwidth}{!}{%
\begin{tabular}{@{}lrrrr@{}}
\toprule
\textbf{Genre}        & \textbf{Docs} & \textbf{Tokens} & \textbf{EDUs} & \textbf{Alignments} \\ \midrule
\textit{academic}     & 2             & 1,952           & 250           & 1,250               \\
\textit{biography}          & 2             & 1,679           & 182           & 910                 \\
\textit{conversation} & 2             & 1,868           & 342           & 1,710               \\
\textit{court}        & 2             & 2,075           & 253           & 1,265               \\
\textit{essay}        & 2             & 2,359           & 301           & 1,505               \\
\textit{fiction}      & 2             & 2,029           & 264           & 1,320               \\
\textit{how-to}       & 2             & 1,642           & 223           & 1,115               \\
\textit{interview}    & 2             & 1,653           & 209           & 1,045               \\
\textit{letter}       & 2             & 1,939           & 224           & 1,120               \\
\textit{news}         & 2             & 1,891           & 200           & 1,000               \\
\textit{podcast}      & 2             & 2,119           & 257           & 1,285               \\
\textit{reddit}       & 2             & 1,858           & 257           & 1,285               \\
\textit{speech}       & 2             & 1,728           & 184           & 920                 \\
\textit{textbook}     & 2             & 2,072           & 255           & 1,275               \\
\textit{travel}       & 2             & 1,722           & 154           & 770                 \\
\textit{vlog}         & 2             & 1,669           & 224           & 1,120               \\
\midrule
\textbf{Total}        & 32            & 30,255          & 3,779         & 18,895              \\ \bottomrule
\end{tabular}
}
\caption{GUM V12 test set data.}
\label{tab:data}
\end{table}

Additionally, the original corpus contain discourse trees following enhanced Rhetorical Structure Theory (eRST, see \citealt{MannThompson1988} and \citealt{zeldes-etal-2025-erst}), which we use below to identify proposition boundaries. Since RST trees can be used to assess the centrality of discourse units based on their nesting depth relative to the tree root, we can evaluate and compare our salience annotations based on the nesting depth data from the trees (see Appendix \ref{sec:appendix-centrality} for details).

\section{Annotation}

Given a text and summaries, our annotation task involves three steps: identifying propositions, aligning them to summaries, and categorizing properties of the alignment.

\paragraph{Defining proposition markables} Although propositional structure can be complex and potentially nested, our goal is to be able to rank spans of text for salience with unambiguous scores, meaning we require a `tiling' approach, in which every word in the document belongs to exactly one proposition. To do this in a principled way, we rely on RST's existing notion of Elementary Discourse Units (EDUs). At a maximum, an EDU can be no larger than a sentence, as in \ref{ex:edu-sent}; this applies even if the sentence is a fragment, in which case we may need to treat a single noun phrase (for example a heading) in the same way as a `proposition', as in example \ref{ex:edu-frag}. However we stress that this is inevitable, since summary content can be triggered by verbless utterances such as headings, which can then reasonably be seen as salient spans. Aside from such fragments, EDUs generally correspond to single-clause propositions, usually with one verb and its arguments \ref{ex:edu-clauses}. Additionally, EDUs may be discontinuous, for example due to interruptions by relative clauses \ref{ex:edu-sameunit}. We use the pre-existing EDU segmentation in the GUM corpus without modifications as markable propositions, and treat discontinuous EDUs as single units.

\ex.$[$\textit{Kim called yesterday.}$]_1$\label{ex:edu-sent}

\ex.$[$\textit{INTRODUCTION}$]_1$\label{ex:edu-frag}

\ex.$[$\textit{They called}$]_1$ $[$\textit{because you forgot it}$]_2$ \label{ex:edu-clauses}

\ex.$[$\textit{The boy}$]_1$ $[$\textit{who laughed}$]_2$ $[$\textit{fell asleep}$]_1$\label{ex:edu-sameunit}

We also considered the alternative of focusing on entire sentences, but realized these would be too coarse grained: sentences in the corpus contain an average of 2.28 EDUs, and our annotations below indicate that while most sentences are not included in summaries at all (about 78.5\%), equal amounts of sentences are mentioned in their entirety (10.7\%) versus partially (10.8\%, in which some EDUs are mentioned and others are not), meaning sub-sentential alignment is common.

\paragraph{Alignment} To decide whether a proposition appears in a summary, we rely on two levels of alignment: a strong alignment match is present if the summary refers to the same event or predicate in the proposition, or for a fragment, refers to the same entity. However in many cases, a proposition is not mentioned exactly, but it is essential for the content of the summary. For example, the summary might mention when the events in a document take place, but not mention the predicate that actually contained the time information. In such cases, we say that the proposition \textbf{triggers} content in the summary, and refer to the alignment as \textbf{approximate}. Our complete guidelines, which give a range of such examples, are made publicly available with this paper.\footnote{\url{https://gucorpling.org/amir/pdf/propsal-guidelines.pdf}.}

\paragraph{Categorization} Proposition alignment proceeds summary-wise, with each document-summary pair receiving a set of binary alignment judgments for each proposition. Once an alignment has been made, we further categorize it into a proposition \textbf{match} versus \textbf{approximate} alignment. For approximate cases, we further sub-categorize alignments into normal triggers, where some information from the proposition is carried into the summary directly, and \textbf{component} alignments, in which a summary mentions an aggregation over propositions in the document. For example, if the summary of an article about where NASA shuttles will be displayed mentions `some people were disappointed with the final location decisions', but no such statement appears in the document directly, the examples in \ref{ex:components} are taken to be components instantiating the triggering information in aggregate.

\ex. \label{ex:components}\a. \textit{While the museum of flight was in the top running, I'm disappointed that NASA did not choose them}
\b. \textit{Cornyn's statement added ... I’m deeply disappointed with the Administration's misguided decision}

Neither of these two propositions appear in the summary directly, but they are the propositions which, taken together, result in the summary's statement about some people's disappointment.

Additionally, if multiple propositions can be made responsible for the same information in the summary (match or approximate), we annotate them as \textbf{duplicates}, indicating that while they provide salient information, there is substantial redundancy between them. Component alignments containing distinct information, as in example \ref{ex:components} are not considered duplicates.

\begin{figure}[h!tb]
\centering
\includegraphics[width=\columnwidth, clip, trim=1.3cm 13.5cm 3cm 1.3cm, frame]{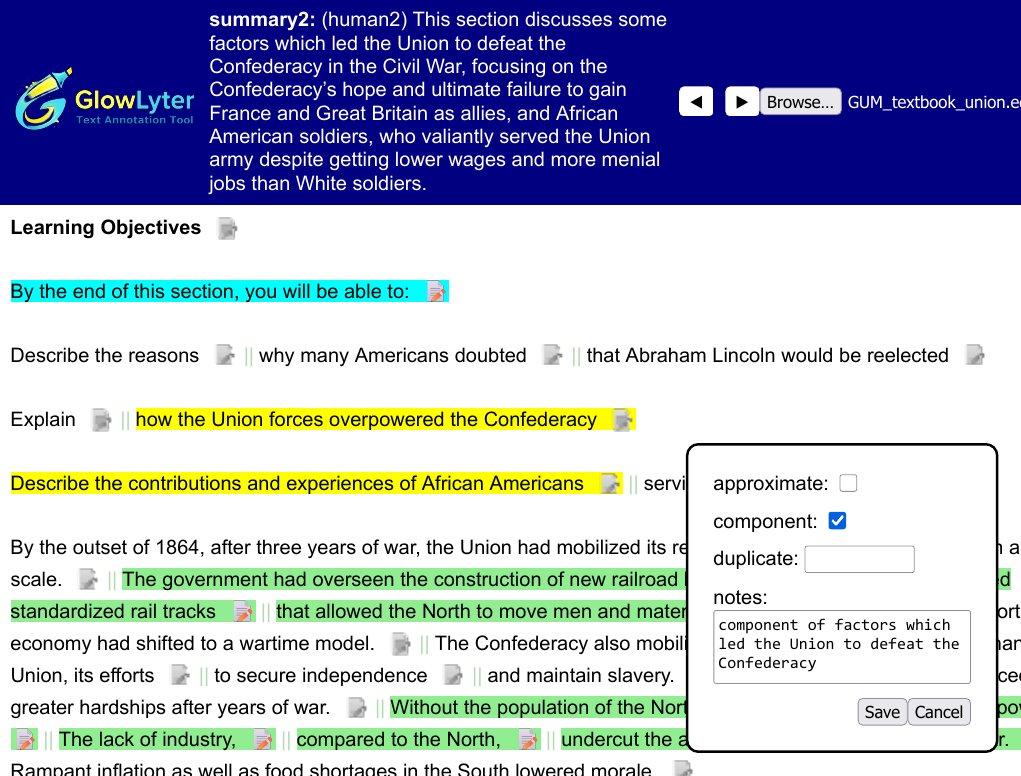}

\caption{GlowLyter annotation interface for summary-wise salient proposition alignment.}
\label{fig:interface}
\end{figure}

\paragraph{Annotation interface} To align propositions we create a dedicated interface called \textbf{GlowLyter}\footnote{\url{https://github.com/gucorpling/glowlyter}} which allows highlighting pre-defined units by clicking, with separate highlighting storage saved for each of the summaries displayed at the top (Figure \ref{fig:interface}). A note button next to each span opens the additional annotation box to indicate approximate, component and duplicate alignment information, which receive unique colors (yellow default highlighting, cyan for `approximate' and green for `component'). The interface can easily be used to produce any kind of highlighting annotations of sentences or other spans and uses a file format that allows definition of custom annotation fields with text boxes or check boxes, making it adaptable to other types of annotations (see Appendix \ref{sec:tool-format}). We release the code for the interface with our data on our GitHub repository under the open Apache 2.0 license. The annotated data will be released as part of the publicly available GUM corpus.

\section{Evaluation}

Because most propositions are not salient ($\sim$90\%, depending on the genre/document), we anticipate that \% agreement (propositions two annotators assign the same label) is an optimistic metric; we therefore also report Cohen's kappa, which takes into account the probability of chance agreement. We measure agreement in four increasingly lenient scenarios: 1. exact agreement, incl. all EDUs;  2. strictly literal agreement, disregarding disagreements involving components; 3. strictly matching agreement, the same but disregarding approximates; and 4. strictly matching duplicate-agnostic agreement, the same as 3. but treating `duplicate' units as interchangeable (annotators agree if they both flag one in a set as salient, or both flag none).

\begin{table}[h!tb]
    \centering
    \resizebox{\columnwidth}{!}{
    \begin{tabular}{lcccc}
    \toprule
        & \multicolumn{2}{c}{accuracy} & \multicolumn{2}{c}{$\kappa$} \\
        metric & micro & macro & micro & macro \\
        \midrule
    \textit{strict all} & 92.97 & 92.73 & 65.43 & 64.62 \\
    \textit{strict literal} & 95.08 & 94.78 & 74.07 & 72.74 \\
    \textit{strict match} & 95.96 & 95.73 & 78.05 & 77.57 \\
    \textit{duplicates OK} & 96.63 & 96.52 & 83.99 & 82.64 \\
    \bottomrule
    \end{tabular}    
    }
    \caption{Inter-annotator agreement}
    \label{tab:agreement}
\end{table}

Table \ref{tab:agreement} shows that strict agreement is only moderate. Tolerating interchangeable duplicates (ways of mentioning `the same thing') substantially raises agreement (+5.9\% micro $\kappa$). Component alignments are responsible for the most disagreements (+8.6\% micro $\kappa$ when disregarded), much more than non-component approximates add (+3.98\%).

Qualitatively we note cases where annotators focus more on lexical overlap than identical reference. For example for a summary mentioning `\textit{a memory of the last time they saw a "yellow man"}', one annotator selected `\textit{I remembered the only "yellow man" I had ever seen}', while another omitted `\textit{I had ever seen}', which references `seeing', but may mean the specific seeing event or experience in general. Fragment component mentions also posed difficulties: one summary mentions a `family', but the text enumerates only the members including two EDUs for the mother: `\textit{my mother twisting her hands}' and `\textit{my father's wife with red eyes}'. Different annotators preferred each of these: one argued for the first based on text order, while the other preferred the second since it did not include a superfluous verb `twisting', a proposition that did not appear in the summary.

\section{Analysis}

\begin{figure}[h!tb]
\centering
\includegraphics[width=0.4\textwidth]{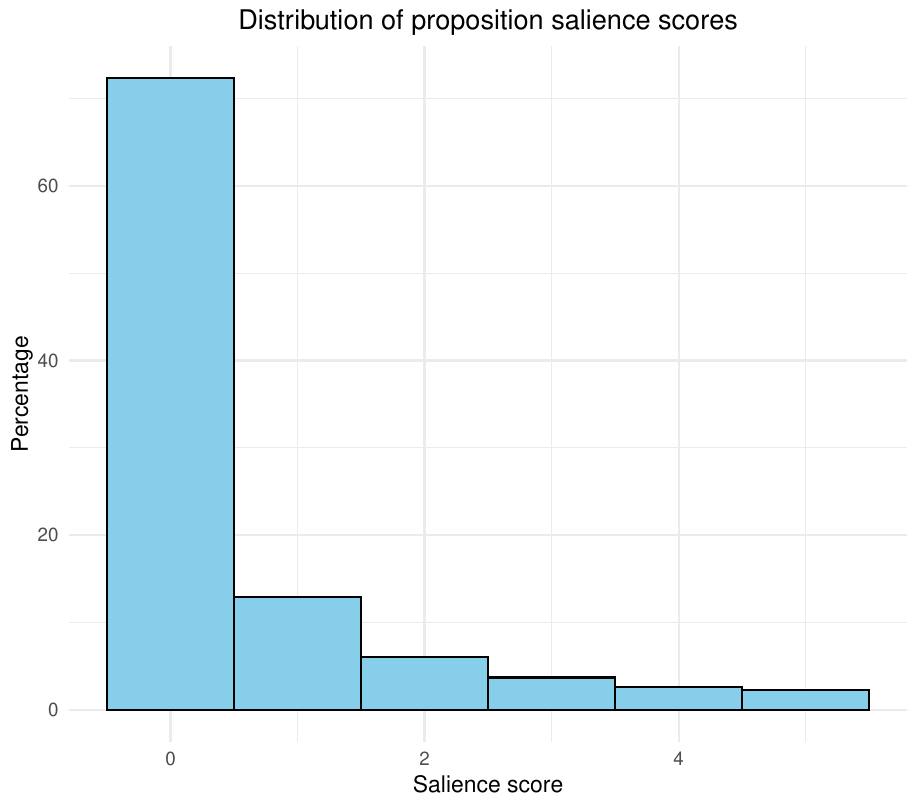}
\caption{Distribution of salience scores.}
\label{fig:histogram}
\end{figure}

Figure \ref{fig:histogram} shows the salience score distribution. Unlike summary-based entity salience scores, which follow a U-shaped distribution \cite{lin-zeldes-2025-gum}, proposition scores follow a descending histogram. Whereas, based on previous work, most entities are not in summaries, but the next biggest group is entities appearing in all summaries, the largest amount of salient propositions are mentioned in only one summary. This suggests inter-summary variance is higher at the level of proposition selection than is the case for participants.

Since we have RST trees for our data, we can also examine how our scores correspond to centrality in RST, i.e. how far an EDU is from the root of the tree. We operationalize this distance as a proportion from 0 (the unit is the root) to 1 (the unit has the maximum distance in its document; see Appendix \ref{sec:appendix-centrality} and Figure \ref{fig:rst-centrality} for details).

We find that RST centrality is significantly correlated  with salience ($r^2=-0.287, p<0.0001$) though the correlation is diminished by many confounds: longer EDUs are more likely to overlap with summaries, as are complete sentences compared to clauses. RST relations, which indicate the discourse role of each unit (expressing \textsc{purpose}, \textsc{cause} etc., see Appendix \ref{sec:appendix-rst-salience}) are also likely to influence salience, as does the EDU's position in the document: the earlier a unit appears, the higher its score ($r^2=-0.119,p<0.0001$). 

In order to test whether RST centrality remains a significant predictor of proposition salience when these confounds are accounted for, we construct a mixed effects model with the document as a random effect. Since our salience scores do not follow a normal distribution (see Figure \ref{fig:histogram}) and only allow for discrete integer values, we must use a target distribution that is compatible with non-normal, and specifically with long-tailed, integer data.\footnote{We thank an anonymous reviewer for commenting on this.} To that end, we use beta-binomial regression, modeling each summary containing a proposition as a single `success', and each proposition as instantiating five trials (chances to be mentioned in a summary). To test the significance of individual features, we build the complete model with all features, and then perform single term deletions, comparing the full model to a model with each feature removed, using a Likelihood Ratio Test (LRT), reported in Table \ref{tab:model}).

\begin{table}[htbp]
\centering
\resizebox{\columnwidth}{!}{
\begin{tabular}{lrrrr}
\toprule
Feature & {df} & {AIC} & {LRT} & {Pr($\chi^2$)}  \\
\midrule

$<$none$>$ & {} & -4000.5 & {} & {} \\
\textit{position} & 1 & -3962.8 & 39.68 & 2.987e-10 *** \\
\textit{is\_sent} & 1 & -3935.5 & 67.02 & 2.682e-16 *** \\
\textit{length} & 1 & -3906.5 & 95.98 & < 2.2e-16 *** \\
\textit{relation} & 31 & -3469.0 & 593.46 & < 2.2e-16 *** \\
\textit{centrality} & 1 & -3198.2 & 804.28 & < 2.2e-16 *** \\

\bottomrule

\end{tabular}
}

\caption{LRT single term deletions for features in a mixed effects beta-binomial regression model predicting salience, ranked by AIC.}\label{tab:model}
\end{table}

The model confirms that centrality is the strongest predictor of salience when confounds are accounted for (maximum difference using Akaike's Information Criterion, AIC). It is followed by the relation type and, with a much weaker effect, by unit length. However the model still achieves only a weak fit to the data (marginal $r^2=0.131$, conditional $r^2=0.159$) and a binary classification accuracy of just 73.87\% over a majority baseline of 72.33\% (always predicting salience=0). This suggests much work remains to be done in understanding what characterizes salient propositions.

\section{Conclusion and outlook}

This paper presented a first effort to manually annotate gradient proposition salience based on alignment of discourse units with multiple summaries, adapting a methodology already in use for entities to the realm of predicates. Our evaluation shows that while the task is challenging, agreement is far above chance, and we plan to apply lessons from our adjudication process to refining our guidelines.

Work on salient entities (see \citealt{zeldes-lin-2026-cllt}) has revealed a range of expected and surprising results about what makes some parts of a text or conversation particularly noteworthy, but much corresponding work on propositions is still pending. The results of the preliminary study above already suggest a range of correlations between the notion of proposition salience developed in this paper and other directly observable properties (document position, length) and theoretical constructs (discourse relations, RST centrality) applying to propositions. Very recent concurrent work on entity salience has also shown correlations with the ways in which text is represented by language models, such as effects related to surprisal \cite{lin-zeldes-2026-expect}, and we are excited by the prospect of discovering additional properties that make propositions more or less salient in a similar vein.

We also intend to annotate the rest of the GUM corpus data and its complementary out-of-domain test set with challenging genres, GENTLE (GENre Tests for Linguistic Evaluation, \citealt{aoyama-etal-2023-gentle}). We plan to apply 
further manual annotation to the dev set and use NLP/LLM-assisted annotation for the train set. The test set created here will be used to evaluate models for the task and develop prompting approaches for LLM ensembles to approach human performance. We hope that this data will enable detailed analyses of what makes propositions salient, and that our publicly released annotation guidelines and interface will be useful for the community.

\section*{Limitations}

This pilot study covers only the test set of one corpus, limited to contemporary English. Although we believe the quality of both the EDU segmentation and the summaries in GUM is high, our annotations are by definition only as good as these underlying sources of information. The error analysis in this short paper is limited to a few examples due to space. We also recognize that agreement on the proposition alignment task is still far from perfect, and are using our experiences from this study to refine guidelines for the final adjudicated data release. We look forward to feedback from the community on ways to make the data as consistent and reproducible as possible.


\bibliography{custom}

\appendix

\section{Annotation tool format}\label{sec:tool-format}

\begin{figure*}
    \centering
    \includegraphics[width=\textwidth, clip, trim=2.3cm 14.5cm 3cm 2.3cm, frame]{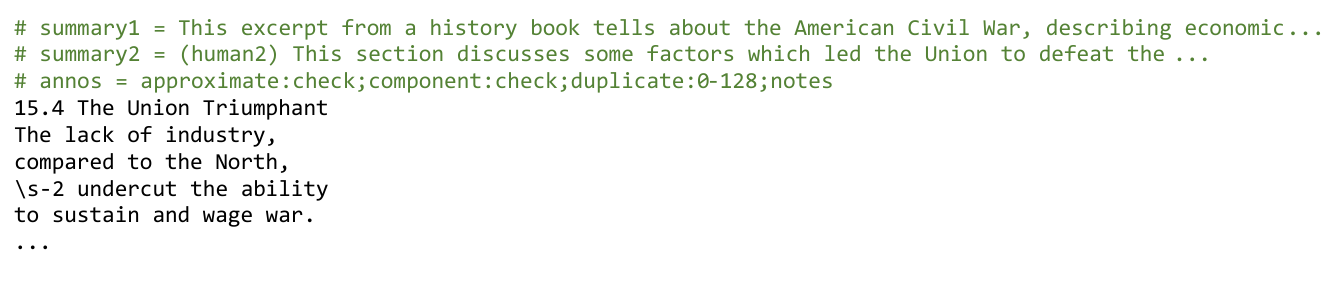}
    \caption{Input format for the annotation interface}
    \label{fig:format}
\end{figure*}

Our annotation tool, GlowLyter, is built for aligning text segments to content displayed in a carousel at the top of the interface, in this case containing five summaries per document. However the tool supports a generic format, shown in Figure \ref{fig:format}, which incorporates a header (lines prefixed by \texttt{\#}) declaring the desired annotation types.

In particular, the \texttt{annos} declaration is responsible for generating the annotation dialog elements seen in Figure \ref{fig:interface}. Check boxes are added using the syntax \texttt{annoname:check}, numerical fields can be added by specifying a range of numbers (here: \texttt{duplicate:0-128}, since this text has 128 units), drop down lists can be added by specifying their values (\texttt{annoname:val1,val2,val3,...}, not used above), and fields with no values are declared as simple key names (in this example, \texttt{notes}). The interface can therefore be used for a variety of annotation tasks.

The spans to be highlighted are given line by line below the header rows. For discontinuous discourse units we use the escape notation {\textbackslash}s-N, where N is the content line that starts the discontinuous span. Thus {\textbackslash}s-2 on the fourth content line in the Figure indicates that this unit (unit 4) is part of the same unit as unit 2, i.e.~the discontinuous span ``\textit{The lack of industry, .. compared to the North,}''. Other special codes can be used, such as {\textbackslash}b to indicate a unit to be displayed in bold (for headings), {\textbackslash}i for italics, {\textbackslash}l-NAME for speaker labels and {\textbackslash}n to add a line of space at the end of paragraphs.

\section{Computing RST centrality}
\label{sec:appendix-centrality}

\begin{figure}[h!tb]
\centering
\includegraphics[width=0.5\textwidth, clip, trim=1.5cm 11.5cm 15.2cm 3.9cm]{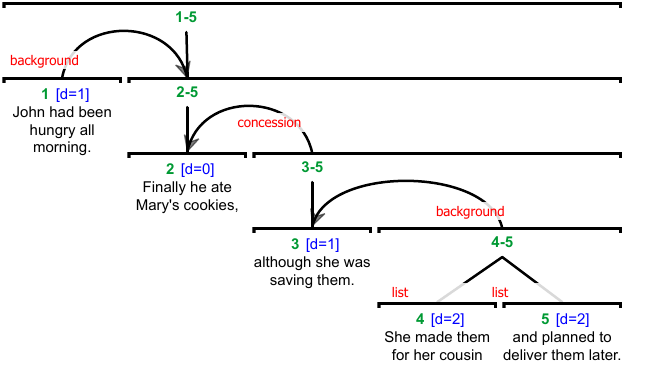}
\caption{RST centrality example. Graph distance from the root is indicated in blue, e.g.~\texttt{$[$d=1$]$} means one horizontal edge away from the root, which is unit $[$2$]$.}
\label{fig:rst-centrality}
\end{figure}

RST trees recursively divide a document into satellite and nucleus units connected by discourse relations. Satellites are considered less prominent than nuclei, and analysts assign satellite status to units that are less central to the pragmatic purpose or main meaning of a document or section. If two units form a coordination of equally important propositions, a multinuclear relation is postulated. For example, in the \textsc{concession} relation in Figure \ref{fig:rst-centrality}, the less important conceded unit $[$3-5$]$ is a satellite of the nucleus, unit $[$2$]$. However Units $[$4$]$ and $[$5$]$ are coordinate and equally important.

The relative centrality of a unit can be measured based on the number of satellite relations (horizontal arrows) that are traversed between it and the document root, that is the unit (or units) which head a constituent that is not a satellite of any other constituent. 

In the Figure, unit $[$2$]$ is the root, which can be understood as the most salient unit from the RST perspective -- John's eating of the cookies is the most central event, and therefore has a distance of 0 (\texttt{$[$d=0$]$}). The constituents headed by $[$1$]$ and $[$3$]$ are direct satellites of the root, and therefore have \texttt{$[$d=1$]$}. Finally the coordination of $[$4$]$ and $[$5$]$ is a satellite of $[$3$]$, and therefore has a greater distance, scoring \texttt{$[$d=2$]$}.

\section{RST vs. salience example}\label{sec:appendix-rst-salience}

\begin{figure*}[b!th]
\centering
\includegraphics[width=\textwidth, clip, trim=1.5cm 4.5cm 3.2cm 3.9cm]{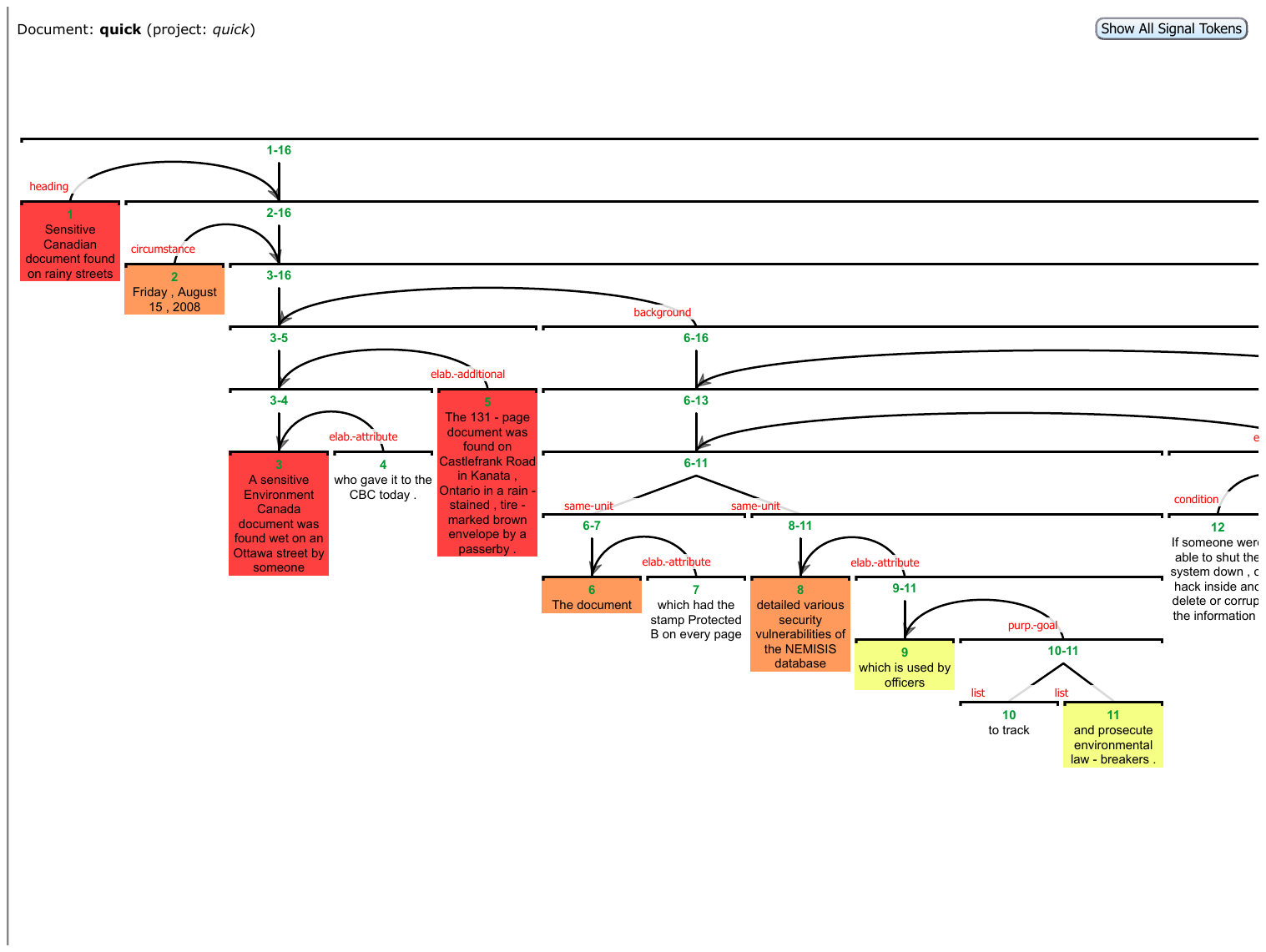}
\caption{Fragment of an RST tree with units shaded by salience: a score of 5=red, 3=orange, 1=yellow.}
\label{fig:salience-shading}
\end{figure*}

Figure \ref{fig:salience-shading} shows a fragment of an RST tree with nodes shaded by their aligned proposition salience. The most central discourse unit $[$3$]$, which all units point to directly or indirectly, receives the maximum score of 5 (shaded red) because all summaries mention this proposition (`\textit{A sensitive .. document was found .. on an Ottawa street}'). This unit is considered a duplicate of the title (unit $[$1$]$), which is as salient, and likewise unit $[$5$]$ receives the same score. The date of the event (`\textit{August 15, 2008}') is mentioned in only 3 summaries, corresponding to the orange highlight for unit $[$2$]$. Since no summary mentions unit $[$4$]$ (the document being given to CBC), it receives no background color and remains white. Unit $[$6$]$ is interrupted by unit $[$7$]$ (not mentioned in any summary), and continued by unit $[$8$]$, marked using the label \textsc{same-unit}. 

Graph topology (graph distance from the most central or root unit, $[$3$]$) is correlated with salience, but not perfectly: for example the coordinate units $[$10$]$ and $[$11$]$ differ: $[$11$]$ is mentioned by one summary (marked in yellow), but $[$10$]$ is not.

\end{document}